\PassOptionsToPackage{table,xcdraw}{xcolor}
\documentclass[manuscript,screen]{acmart}
\usepackage{longtable}
 \usepackage{booktabs}
 \usepackage{graphicx}

\usepackage{amsmath,amssymb,amsfonts}

\usepackage{algorithmic}
\usepackage{textcomp}

 \usepackage{multirow}
 \usepackage[table,xcdraw]{xcolor}
 \usepackage{dblfloatfix}
\usepackage{bigstrut}
\usepackage{multirow}
\usepackage{color}
\usepackage{pgfplots}
\usepackage{balance}
\usepackage{enumitem}

\AtBeginDocument{%
  \providecommand\BibTeX{{%
    \normalfont B\kern-0.5em{\scshape i\kern-0.25em b}\kern-0.8em\TeX}}}

\setcopyright{acmcopyright}
\copyrightyear{2018}
\acmYear{2018}
\acmDOI{XXXXXXX.XXXXXXX}

\acmConference[Conference acronym 'XX]{Make sure to enter the correct
  conference title from your rights confirmation emai}{June 03--05,
  2018}{Woodstock, NY}
\acmPrice{15.00}
\acmISBN{978-1-4503-XXXX-X/18/06}

\begin{document}

\title{Deep Learning based Fingerprint Presentation Attack Detection: A Comprehensive Survey}

\author{Hailin Li}
\email{hailin.li@ntnu.no}
\orcid{1234-5678-9012}
\author{Raghavendra Ramachandra}
\orcid{0000-0003-0484-3956}
\email{raghavendra.ramachandra@ntnu.no} 
\affiliation{%
  \institution{Institute of Information Security and Communication Technology (IIK), Norwegian University of Science and Technology (NTNU)}
  \city{Gj\o{}vik}
  \country{Norway}
  \postcode{2816}
}








\renewcommand{\shortauthors}{Hailin and Raghavendra}

\begin{abstract}
The vulnerabilities of fingerprint authentication systems have raised security concerns when adapting them to highly secure access-control applications. Therefore, Fingerprint Presentation Attack Detection (FPAD) methods are essential for ensuring reliable fingerprint authentication. Owing to the lack of generation capacity of traditional handcrafted based approaches, deep learning-based FPAD has become mainstream and has achieved remarkable performance in the past decade. Existing reviews have focused more on hand-cratfed rather than deep learning-based methods, which are outdated. To stimulate future research, we will concentrate only on recent deep-learning-based FPAD methods. In this paper, we first briefly introduce the most common Presentation Attack Instruments (PAIs) and publicly available fingerprint Presentation Attack (PA) datasets. We then describe the existing deep-learning FPAD by categorizing them into contact, contactless, and smartphone-based approaches. Finally, we conclude the paper by discussing the open challenges at the current stage and emphasizing the potential future perspective.

\end{abstract}



\received{20 February 2007}
\received[revised]{12 March 2009}
\received[accepted]{5 June 2009}


\maketitle

\section{Introduction}
\label{sec:introduction}

Biometric verification is widely deployed in numerous access control applications, including border control, forensic science, smartphone access, attendance systems, etc. Biometric systems can be designed using either physiological (e.g. fingerprint, face, iris), behavioral (e.g. gait, keystroke, voice) or a combination of both. Among various physiological traits, the face, iris, and fingerprint have dominated the majority of applications because of their reliability and accuracy in performance, which can be attributed to the uniqueness of these biometric characteristics. However, the fingerprint biometrics is one of the traditional biometric characteristics in various application considering the reliability of finger patterns over long period of time and the representation and matching of the features that can achieve  billion fingerprint comparison in a single second with high accuracy \cite{AFIS}.

\begin{figure}[htp]
\begin{center}
\includegraphics[width=1.0\linewidth]{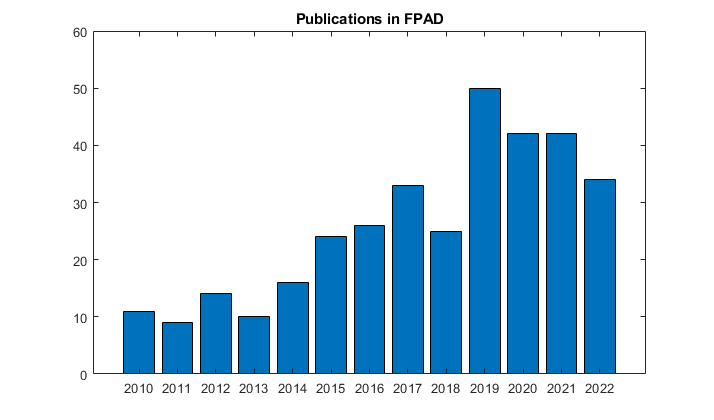}
\caption{Publications of FPAD in recent years obtained through Google scholar search with keywords: “fingerprint spoof detection”, “fingerprint presentation attack detection”, and “fingerprint liveness detection”.}
\label{fig:Survey}
\end{center}
\end{figure}

Fingerprint images are captured using contact and contactless sensing among which the contact-based sensing is used predominantly. Contact-based fingerprint sensing includes optical,  capacitive, and ultrasonic sensors that are available as standalone devices and/or integrated devices in smartphones. Optical fingerprint scanners are the oldest type of capturing fingerprint patterns. These sensors typically have a very high number of diodes per inch to capture the fine details of the finger and the optical camera comes with a finite resolution. Capacitive sensors use arrays of small capacitor circuits to capture the fingerprint data. Since capacitors store electrical charges, connecting them to conductive plates on the scanner's surface allows them to track the details of a fingerprint. Recently, ultrasonic sensors have been introduced to scan fingerprints, particularly in smartphones. These sensors employ an ultrasonic transmitter that can send the pulse against the finger and record the echoes used to construct the fingerprint. Further, Scanning for extended periods allows for additional depth data to be captured, resulting in a 3D fingerprint. There are two types of contactless sensing: (a) use of custom industrial cameras (b) off-the-shelf smartphone cameras used to capture finger photos. The use of custom cameras for contactless capture will allow designers to include multi-spectral cameras and other sophisticated cameras to ensure reliable verification with resilience  to presentation attacks. However, fingerprint capture using contactless sensing introduces additional challenges, such as uncontrolled poses, low-quality fingerprints, environmental noises, and degraded performance.

The widespread deployment of the Fingerprint Recognition System (FRS) has raised concerns about the system being attacked, and the attacker may maliciously gain access to the fingerprint system. FRS can be attacked in two ways (a) direct attack and (b) indirect attacks. Direct attacks typically target the sensors of the FRS such that a Presentation Attack Instrument (PAI) is presented to the sensor to gain access. An indirect attack aims to attack the biometric subsystem components to modify the functionality. Compared to direct attacks, indirect attacks require special skills and knowledge of the biometric system to successfully gain access to the FRS. Therefore, direct attacks on FRS have been used extensively in real-life scenarios. One real-life example of hacking the FRS using direct attacks, mainly on the national population registry, was reported in \cite{HackingAdhaar}. The Aadhaar-enabled Payment System (AePS) is spoofed to withdraw money from various victims. Attackers achieved this by collecting the fingerprints of the victims from registry papers that have ink-prints of fingerprints. The attackers then created a polymer fingerprint attack instrument that was used to withdraw money through the AePS. Therefore, the detection of presentation attacks is of paramount importance in ensuring the security of the FRS system to facilitate reliable verification.


\begin{table*}[htbp]
\centering

\resizebox{\columnwidth}{!}{%
\begin{tabular}{p{20.355em}|c|c|p{12.355em}}
\hline
\textbf{Paper Title / Reference} &\textbf{ Year} & \textbf{Deep Learning included} & \textbf{Modality \& Hardware}  \\
 \hline
Survey on fingerprint liveness detection \cite{al2013survey} &2013 & No & Contact based\\
\hline
 Presentation attack detection methods for fingerprint
recognition systems: a survey \cite{sousedik2014presentation} & 2014& No&Contact based\\
\hline
 A Survey on Antispoofing Schemes for Fingerprint Recognition
Systems \cite{marasco2014survey} & 2014 & No & Contact based  \\
\hline
Survey on Fingerprint Spoofing, Detection Techniques and Databases\cite{kulkarni2015survey} & 2015 & No & Contact based \\
\hline
Security and Accuracy of Fingerprint-Based
Biometrics: A Review \cite{yang2019security} & 2019 & Few & Contact based \\
\hline
A Survey on Unknown Presentation Attack Detection
for Fingerprint \cite{singh2021survey} & 2021 & Few & Contact based, SWIR, LSCI \\
\hline
Robust anti-spoofing techniques for fingerprint
liveness detection: A Survey \cite{habib2021robust} & 2021 &Few& Contact based \\
\hline
FinPAD: State-of-the-art of fingerprint presentation attack detection
mechanisms, taxonomy and future perspectives \cite{sharma2021finpad}& 2021 & Yes,<30 &Contact based,  SWIR\\
\hline
Fingerprint Liveness Detection Schemes: A Review
on Presentation Attack \cite{ametefe2022fingerprint} & 2022 &Yes, <30 & Contact based, SWIR, LSCI, smartphone \\
\hline
\textbf{Deep Learning for Fingerprint Presentation Attack Detection: A Survey (Ours)} & 2023 & Comprehensive (>50) & Contact based, SWIR, LSCI, FTIR, OCT, smartphone \\
\hline
\end{tabular}
}
\caption{A summary of existing surveys in FPAD}
\label{tab:Existing survey}
\end{table*}

Presentation Attack Detection (PAD), a.k.a anti-spoofing method, has been widely investigated for fingerprint biometrics, resulting in several PAD techniques.  The progress of Fingerprint PAD (FPAD) techniques is shown in Figure \ref{fig:Survey} which illustrates the increased interest of researchers in developing FPAD techniques. Early works on developing FPAD  techniques were based on hand-crafted features, in which texture-based features (Local Binary Patterns (LBP), Binarized Statistical Image Features (BSIF), etc.) were widely employed in designing FPAD. However, owing to the limitations of the texture features to generalize across different types of PAIs, researchers have started to investigate FPAD using deep learning.  Further, the importance of FPAD also resulted in the development of the competition platform LivDet \cite{LivDetWebpage} which allows participants to submit FPAD algorithms for independent evaluation.

The exponential growth of the FPAD algorithms over the years has resulted in several survey papers, as listed in Table \ref{tab:Existing survey}. However, existing surveys are limited to non-deep learning approaches and thus do not consider the recent progress in which deep learning approaches are prominent.  Hence, in this paper, we are motivated to present a comprehensive review of deep learning-based fingerprint presentation attack detection, in which we discuss recent progress, competition, performance evaluation metrics, and future work. The main contributions of this study are as follows:

\begin{itemize}
    \item We present the comprehensive survey on deep-learning based FPAD techniques for both contact and contact less fingerprint. Comprehensive literature survey together with the taxonomy and compare these approaches on  the different attributes of design. 
    \item We present a comprehensive survey on the PAI that are widely employed in both contact and contactless fingerprint biometrics. 
    \item We outline the main challenges and the potential future work for reliable fingerprint detection.  
\end{itemize}

The rest of the paper is organized as follows. Section \ref{sec:FRS} presents the pipeline of a fingerprint recognition system and indicates how the FPAD system can be adjusted in the overall system. Section \ref{sec:FPSD} provides a  comprehensive report on the different types of PAIs for both contact and contactless fingerprints. Section \ref{sec:Dataset} introduces the most common publicly available FPAD dataset. Section \ref{sec:FPAD} presents a comprehensive survey of fingerprint presentation attack detection based on deep learning. Then, Section \ref{sec:Evaluation} includes the most common PAD performance evaluation metrics from the ISO standard and the LivDet competition. Section \ref{sec:Future} discusses the open challenges and the potential future research directions. Finally, we conclude this review in section \ref{sec:Conclusion}.

\section{Fingerprint Recognition Systems (FRS)}
\label{sec:FRS}
\begin{figure*}[htbp!]
\begin{center}
\includegraphics[width=1.0\linewidth]{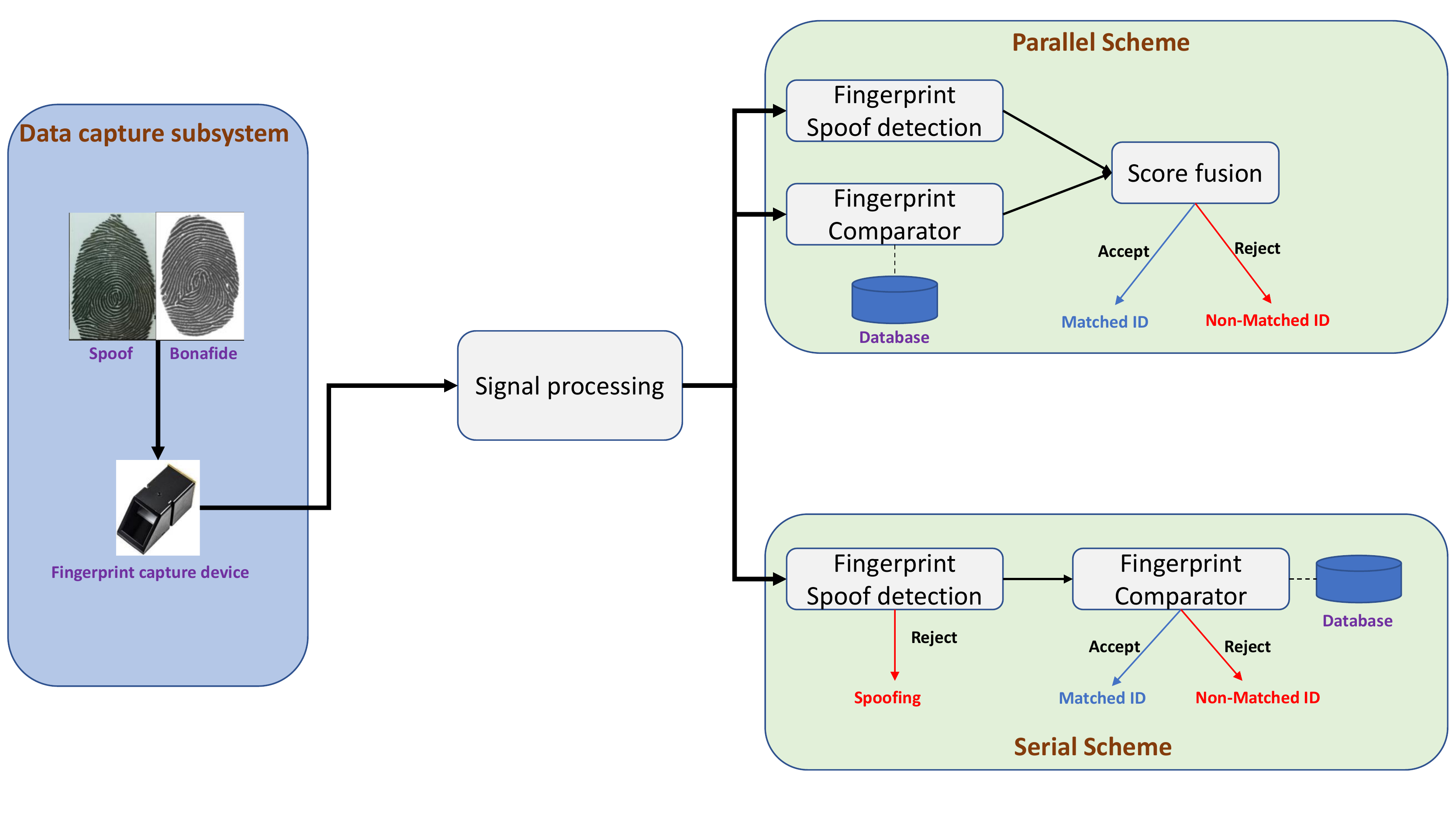}
\caption{Block diagram of fingerprint verification system with PAD}
\label{fig:pipeline}
\end{center}
\end{figure*}
Figure \ref{fig:pipeline} shows the block diagram of the fingerprint verification system. Given a fingerprint image, the signal processing unit performs various operations, including data preprocessing, quality checking, feature extraction, and template creation. Fingerprint feature extraction techniques can be  level 1 (pattern), level 2 (minutiae points), and level 3 (pores and ridge shape).  The common and successful fingerprint features that are widely deployed in commercial applications are based on Level 2 features. Final decisions are made using a comparator that can compare the enrolled fingerprint with the probe fingerprint image to output the comparison score. The comparison score was then compared with the preset threshold to make the final decision.  The PAD system can be integrated into a fingerprint recognition system, either parallel or serial to the fingerprint comparator. In a parallel system, both the PAD and comparator perform processing independently to obtain the decisions that are combined to make the final decision. However, in a serial system, the PAD and fingerprint comparator work in a sequential manner in which the fingerprint is first processed through the PAD unit. If the output of the PAD unit indicates a bona fide, then the fingerprint template is passed through the comparator to make the final decision. For more detailed information on the fingerprint systems,  readers can refer to \cite{maltoni2009handbook} \cite{maltoni2022handbook}.

\section{Fingerprint Presentation Attack Instrument (PAI)}
\label{sec:FPSD}

\begin{figure*}[htbp!]
\begin{center}
\includegraphics[width=1.0\linewidth]{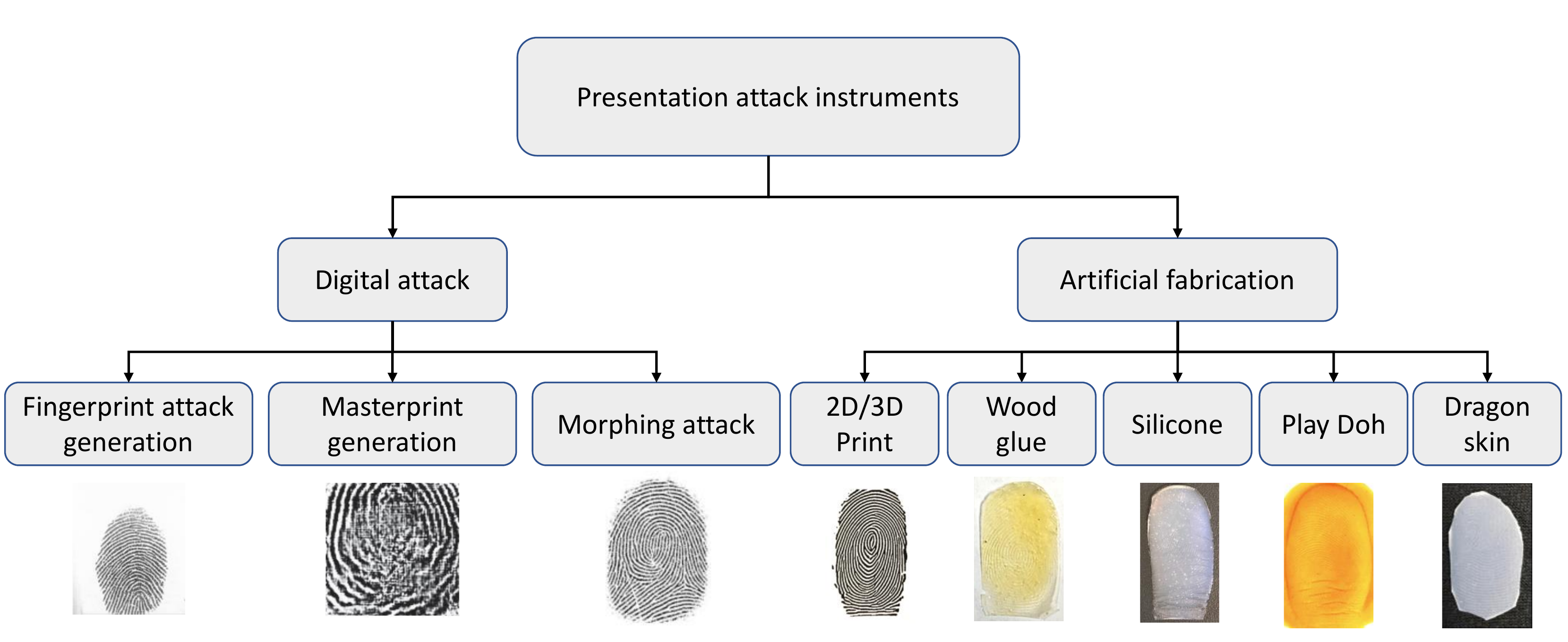}
\caption{Taxonomy of fingerprint Presentation Attack Instrument (PAI)}
\label{fig:PAI_taxonomy}
\end{center}
\end{figure*}

The success of a presentation attack on fingerprint systems depends on the high-quality generation of PAI. Figure \ref{fig:PAI_taxonomy} shows the taxonomy of the existing PAI types commonly studied in the fingerprint literature \cite{maltoni2009handbook}. Available PAI can be broadly categorized into two main types: (a) digital generation and (b) artificial fabrication. Digital attack generation is performed using a computer program in which the attacks are synthetically generated using deep learning methods \cite{karras2019style} or the custom algorithms \cite{bontrager2018deepmasterprints} \cite{gajawada2019universal} \cite{engelsma2022printsgan} \cite{kim2019fingerprint}. Digital attacks can be used as injection attacks and/or to generate physical artifacts that can be used as presentation attacks. Examples of digital attacks include synthetic fingerprint attacks \cite{grosz2022spoofgan}, master print \cite{roy2018evolutionary} and morphing attacks \cite{makrushin2021feasibility} \cite{ferrara2016feasibility}.  The artificial fabrication method uses the target fingerprint impression top to generate a physical artifact that can be used as a presentation attack. Artificial fabrication methods can be broadly categorized into 2D/3D printing and gummy fingers. The  2D/3D print,  in which the physical artifacts are printed fingerprints, can be used as presentation attacks \cite{espinoza2011vulnerabilities} \cite{kanich2018use} \cite{prabakaran2020synthesis}.  The gummy fingerprints are made of various materials (Gelatin, Play-doh, Silicone, etc.) that may contain a specific fingerprint template and have shown potential risks to the FRS \cite{chugh2019fingerprint}. Table \ref{tab:PAIcomp} lists the different features  of   the Digital  and Artificial fabrication types of PAI generation.

\begin{table}[htp]
\centering
\resizebox{\columnwidth}{!}{%
\begin{tabular}{@{}|c|c|@{}}
\toprule
\rowcolor[HTML]{EFEFEF} 
\textbf{Digital   PAI}                     & \textbf{Artificial   Fabrication PAI}                     \\ \midrule
Generate high-quality   attack instrument      & Generate near   high-quality attack instrument                \\ \midrule
High attack potential                          & Moderate   attack potential                                   \\ \midrule
Able to   attack multiple identities in single & Mostly designed   to attack single identity                   \\ \midrule
Requires more   technical knowledge            & No need of more   technical knowledge                         \\ \midrule
High-computation   cost                        & Low   computation cost                                        \\ \midrule
Low-cost   generation                          & High-cost   generation                                        \\ \midrule
Very   challenging to detect                   & Easy to detect   particularly with the multi-spectral sensors \\ \bottomrule
\end{tabular}%
}
\caption{Comparison of different PAI generation techniques}
\label{tab:PAIcomp}
\end{table}

\begin{figure*}[htbp!]
\begin{center}
\includegraphics[width=0.8\linewidth]{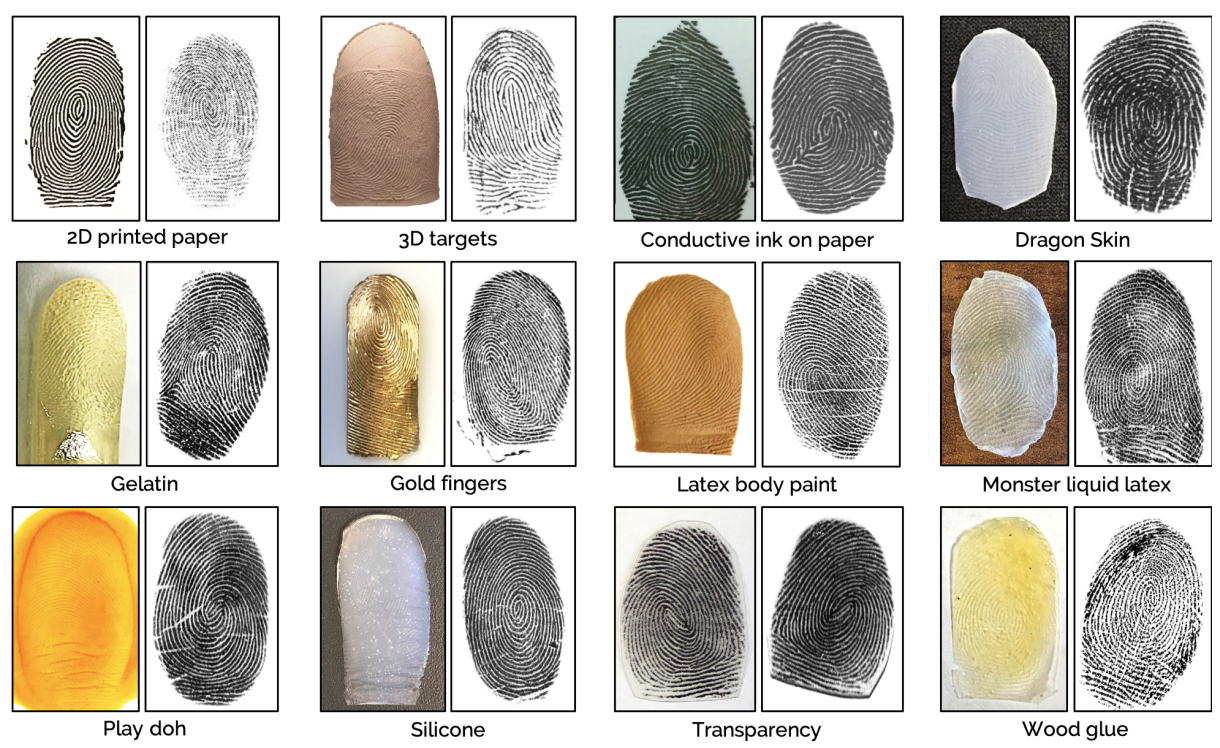}
\caption{Examples of PAIs (taken from \cite{chugh2019fingerprint})}
\label{fig:PAI}
\end{center}
\end{figure*}
The common PAI examples are shown in figure \ref{fig:PAI}. In accordance with the experiments under the fingerprint verification application, the digital synthetic fingerprint shows the high attack potential against the FRS.

\section{Existing Datasets for Fingerprint PAD}
\label{sec:Dataset}
A large-scale dataset is required to achieve better results when using deep-learning-based methods during both the training and testing phases. In this section, we summarize the publicly available FPAD datasets containing the data amount, subject numbers, bona fide/attack amount, and used PAI species. The most common datasets were from the Fingerprint Liveness Detection Competition (LivDet) series from 2009 to 2021. Readers can refer to a review of the LivDet series \cite{micheletto2022review} for detailed information on the LivDet challenge. In the first edition of the LivDet 2009 dataset, three optical scanners, Crossmatch, Identix, and Biometrika, were included with gelatin, silicone, and play-doh as spoof materials. In LivDet 2011, four sub-datasets were based on four different optical scanners: Biometrika, Digital Persona, ItalData, and Sagem. In addition, more PAIs such as Silgum, Ecoflex, Wood Glue, and Latex are included than LivDet 2009. The first non-optical scanner was introduced in the competition, in which the spoof materials were replenished with body doubling and modasil. In LivDet 2015, with the raised concern of FPAD methods against unseen attacks, the competition included some unknown materials for evaluation in the test dataset. Additionally, liquid Ecoflex and a two-component silicone rubber(RTV) are included in the Green Bit, Biometrika, and Digital Persona datasets, whereas the silicone rubber OOMOO is included in the Crossmatch dataset. In LivDet 2017, the competition focused on the impact of the FPAD based on user-specific effects and operator skills in fabricating replicas. In the training set, the spoofing is made of wood glue, Ecoflex, and body double, whereas gelatin, latex, and liquid Ecoflex compose the test set in order to stimulate the completely unseen scenario. Furthermore, two sets of people with different manufacturing spoofing abilities were involved in spoofing. In contrast to the previous editions, some subjects were included in the training and testing datasets to explore the impact of user-specific effects. LivDet 2019 utilized the same scanners as in the previous edition, but presented multi-material combinations, both of different consistency and of different nature for the first time. LivDet 2021 only consists of two scanners, GreenBit and Dermalog, in which the consensual approach and the new pseudo-consensual method, ScreenSpoof, are included. Novel materials RProFast, Elmers glue, gls20, and RPro30.  were chosen for the fakes. In the most recent LivDet 2023, four datasets correspond to two known sensors, Green Bit and Dermalog, and two unknown capture devices.

Besides LivDet datasets, there are also many proposed methods that include custom datasets which is publicly available. In Tsinghua dataset \cite{jia2007new}, the Capacitive device Veridicom is utilized as the capture device and the attack samples are created using Silicone. The samples are acquired from 15 volunteers from Tsinghua University. Two fingers are captured for each participant and ten image sequences
are recorded for each real finger. 47 fake finger are manufactured with ten image sequences
for each finger. BSL dataset \cite{antonelli2006fake} are collected at the Biometric System Laboratory of the University of Bologna which contains more subjects and samples using four different PAIs. For each of 45 volunteers, the thumb and forefinger of the right hand with 10 image sequences are recorded. Instead of making whole 3D fingers, the obtained manufactures are focused on the fingertip area. 
ATVS dataset utilize slilicone and play-Doh as materials and involved 17 subjects. Two scanners are used in Precise Biometrics dataset comprised of 100 subjects and 500 attack samples produced using five different materials. In Precise Biometrics Spoof-Kit(PBSKD) \cite{chugh2018fingerprint}, 900 spoof fingerprint images are fabricated using 10 different types of spoof materials. Another dataset MSU-FPAD is also proposed in \cite{chugh2018fingerprint}, there are up to 9000 live samples and 10500 spoof samples included using these two readers and 4 different spoof fabrication materials. Recently, Kolberg et.al \cite{kolberg2023colfispoof} publish a new dataset COLFISPOOF based on contactless fingerphoto which contains 72 different PAI species. All the fingerprints for the PAIs are generated using fingerprint synthetic algorithms. The details of each dataset are demonstrated in Table \ref{tab:datasets}.

\begin{table*}[htbp]
\centering
\resizebox{\columnwidth}{!}{%
\begin{tabular}{c|c|c|c|c}
 \hline
 \textbf{Dataset}& \textbf{No. of Subjects} & \textbf{Bona fide samples} & \textbf{Attack samples} & \textbf{PAI type}\\
 \hline \hline
Tsinghua \cite{jia2007new}& 15 & 300 & 470  &  Silicone\\
 \hline
BSL \cite{antonelli2006fake}& 45 &900  & 400 & Silicone, gelatin, latex, wood glue\\
 \hline
LivDet 2009 \cite{marcialis2009first} &254  & 5500 &5500  &Gelatine, Silicone and Play-Doh \\
 \hline
LivDet 2011 \cite{yambay2012livdet}  & 200 & 3000 &3000  & Gelatine, Silgum, Ecoflex, Wood Glue, Play-Doh, Silicone and Latex\\
 \hline
LivDet 2013 \cite{ghiani2013livdet}& 225 &  8000&8000  &Gelatine, Wood Glue, Latex, Ecoflex and Modasil \\
 \hline
LivDet 2015 \cite{livDet2015}& 100 &  4500&  5948& Body Double, EcoFlex, Play-Doh, Gelatine, Latex, Wood Glue and Liquid Ecoflex \\
 \hline

LivDet 2017 \cite{livdet2017}& 150 &  8099& 9685 & Gelatine, Wood Glue, Latex, Ecoflex, Body Double and Liquid Ecoflex\\
 \hline
LivDet 2019 \cite{orru2019livdet}&  NA&  6029& 6936 & Gelatine, Wood Glue, Latex, Ecoflex, Body Double and Liquid Ecoflex\\
 \hline
LivDet 2021 \cite{livdet2021}&66  &10700  & 11740 & GLS20, Body Double, Mix 1, ElmersGlue, and RFast30 \\
 \hline
 LivDet 2023 &25  &5000  & 3000 & NA\\
 \hline
 ATVS-FFp \cite{galbally2011evaluation} & 17 & 816 & 816 &Silicone, Play-Doh\\
 \hline
  Precise Biometrics Spoof-Kit \cite{chugh2018fingerprint}& NA & 1000 &900  & Ecoflex, Gelatine, Latex, Crayola, Wood glue,  2D print\\
  \hline

 MSU-FPAD \cite{chugh2018fingerprint}& NA &9000  & 10500 & Ecoflex, 2D Print-Matte Paper, Play-Doh and 2D Print (Transparency)\\
 \hline
 COLFISPOOF \cite{kolberg2023colfispoof} & NA &NA  & 7200 & Print out and Replay\\
 \hline
\end{tabular}
}
\caption{Most utilized and publicly available datasets}
\label{tab:datasets}
\end{table*}

\section{Deep learning based Fingerprint presentation attack detection}
\label{sec:FPAD}
In this section, we discuss the Fingerprint PAD (FPAD) algorithms that are presented for contact, contactless, and smartphone-based fingerprint sensing. The FPAD aims to detect whether the given fingerprint image is bona fide or a presentation attack. FPAD techniques can be broadly classified into two main categories (a) Hardware-based approaches and (b) Software-based approaches. The hardware-based approaches are designed to extract the liveness cues that require explicit (or dedicated) sensors to be integrated into the conventional contact fingerprint biometric system.  Some of the widely used liveness measures include the capture of blood flow \cite{drahansky2006liveness}, Electro-tactile \cite{yau2008fake}, and pulse Oximetry \cite{reddy2008new} to detect whether the fingerprint is live or not. Over the past few years, new expensive sensors like optical coherence tomography (OCT) have evolved. OCT is an imaging technique that allows some of the subsurface characteristics of the skin to be imaged and extracts relevant features of multi-layered tissues up to a maximum depth of 3 mm \cite{cheng2007vivo} \cite{bossen2010internal} \cite{liu2010biometric}. Further, several contactless fingerprint capture devices like multi-spectral and 3D capture devices can also inherently capture the signature of liveness.

The Software-based approach refers to algorithms to detect whether the presented fingerprint is bona fide or a presentation attack, irrespective of the capture device. Software-based FPAD can be further divided broadly into two types handcrafted-based and deep learning. Handcrafted-based method refers to the conventional feature representation that includes techniques used to extract features like gradients, texture, micro-textures, etc. The handcrafted-based approach is applied to contact-based fingerprint images mostly. From the fingerprint images, micro-textural features that can be computed using Scale-invariant feature transform (SIFT) \cite{SIFT}, Binarized Statistical Image Features (BSIF) \cite{BSIF}, Local Binary Pattern (LBP) \cite{LBP},  Local Phase Quantization (LPQ) \cite{LPQ}. Many hand-crafted based approaches like \cite{ghiani2012fingerprint} \cite{ghiani2013fingerprint} \cite{gragnaniello2013fingerprint} \cite{zhang2014fake} \cite{gragnaniello2015local} achieved promised performance in recent years. However, handcrafted-based approaches may have a limitation in that the extraction process becomes difficult due to variations in the acquired fingerprint image quality. These challenges are tackled using Deep Neural Network (DNN) terms such as deep learning, which hierarchically learns deep-level features from the images. With the rapid development of graphical processing units training a large-scale model has become achievable. In 2012, Krizhevsky et al. \cite{NIPS2012_4824} trained a network to classify the 1.2 million high-resolution images in the ImageNet LSVRC-2010 contest into 1000 different classes which achieved a huge success, which started a revolution in the computer vision field. Afterward, training a deep convolutional neural network (CNN) has dominated the image classification tasks in various applications.
Figure \ref{fig:DL category} shows the taxonomy of the deep learning based algorithms that are developed for fingerprint PAD on three different types of sensing that are discussed below: 


\begin{figure*}[htbp!]
\begin{center}
\includegraphics[width=1.0\linewidth]{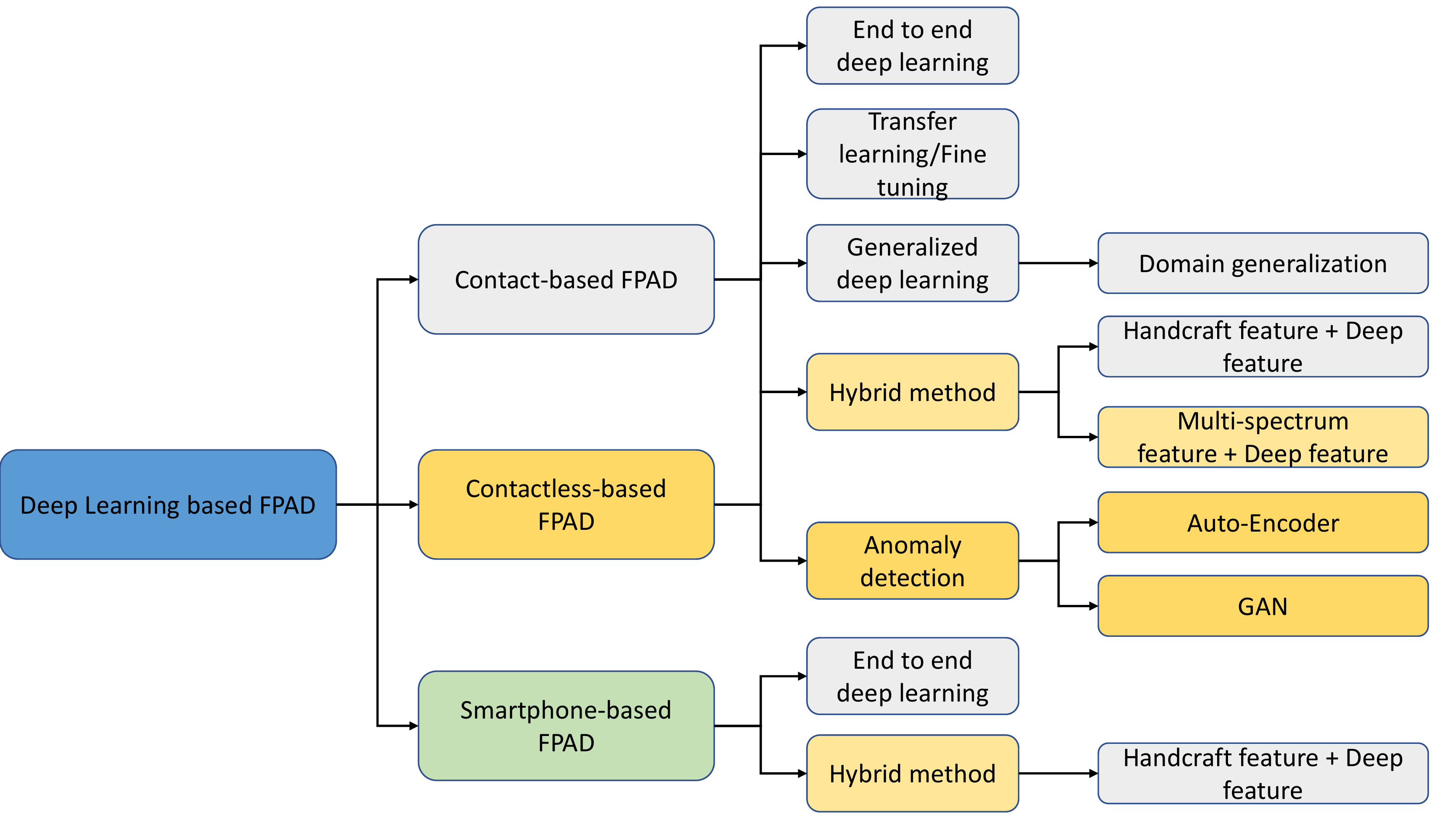}
\end{center}
\caption{Taxonomy of deep learning based fingerprint presentation attack detection}
\label{fig:DL category}
\end{figure*}

\subsection{Contact based FPAD}
Benefited from the rapid development of robust CNN architectures \cite{VGG} \cite{ResNet} \cite{DenseNet} as well as the advanced regularization techniques \cite{srivastava2014dropout} \cite{ioffe2015batch}, researchers have paid more attention to employing deep neural networks to detect fingerprint attacks reliably. Instead of extracting texture features through handcrafted-based descriptors, deep-learning-based approaches can learn deep features that directly map fingerprint inputs to spoof detection. The traditional CNN architecture comprises convolutional layers and a pooling layer that convolves several filters to map the input images to deep-learnable features. The extracted features can be further fed into a fully connected layer for classification tasks. As shown in Figure \ref{fig:DL category}, deep-learning-based methods can be generally divided into two categories. Supervised learning is a straightforward way to determine bona fide and PA as a binary classification task. However, these approaches may not be generalizable to unseen domain attacks (i.e., unknown presentation attacks). Many researchers have considered generalized deep-learning models that can achieve domain generalization to enhance the generalization capacity against unseen PA types.

Initially, Nogueira et al. \cite{nogueira2014evaluating} first introduced a conventional network for fingerprint feature extraction. They trained a Support Vector Machine (SVM) classifier to detect presentation attacks based on CNN deep features and LBP features. To improve the classifier’s performance, several data augmentation techniques such as frequency filtering, contrast equalization, and Region Of Interest (ROI) filtering have been applied. Through comparison, the experiments indicate
high classification accuracy of CNN model which leads to a new direction that inspires more researchers to concentrate on using Deep learning based approaches on FPAD tasks. However, the feature extraction and classification tasks are designed into two different parts so that the model is not optimized simultaneously.

\begin{figure*}[htbp!]
\begin{center}
\includegraphics[width=1.0\linewidth]{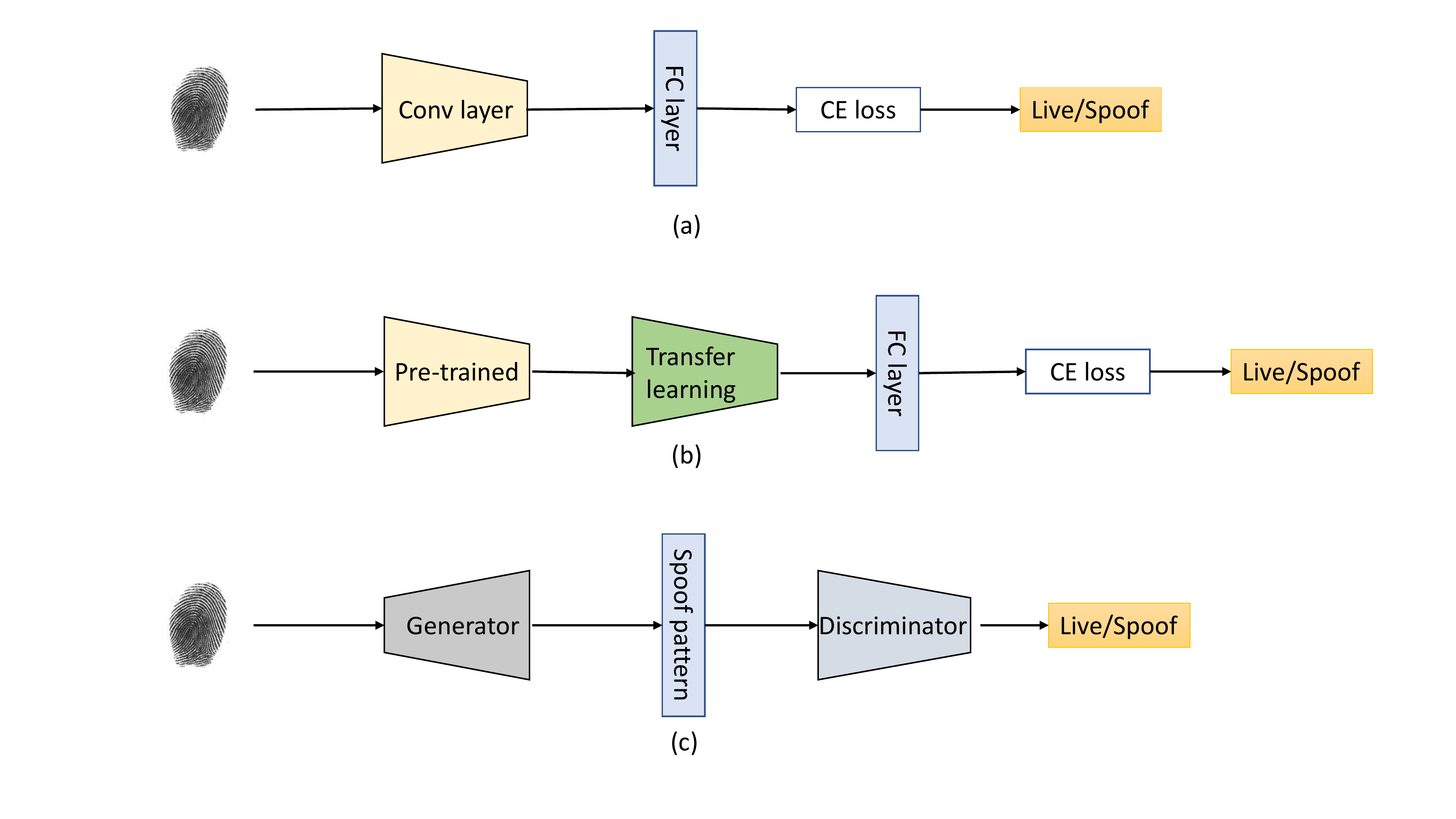}
\end{center}

\caption{Deep learning frameworks for Contact based FPAD. (a) End to end deep learning model using cross-entropy loss. (b) Transfer learning/fine tuning based FPAD approach. (c) FPAD using generalized deep learning model}
\label{fig:DL contact}
\end{figure*}

\subsubsection{End-to-end deep learning}
As shown in figure \ref{fig:DL contact}, end-to-end deep learning models were trained to automatically outperform classification tasks. The feature representation of the fingerprint image was extracted from the convolutional layer and passed into the fully connected layer to calculate the liveness probability using the softmax function.  Table \ref{tab:end to end DL} presents a short description of existing end-to-end deep learning techniques.

In 2015, Wang et al. \cite{wang2015dcnn} divide the input labeled fingerprint images into $32 \times 32$ pixel nonoverlapped patches and pass them into the CNN model for training. Then, the authors adopt a voting strategy to integrate the labels of all the patches to finally determine the result. Similarly, Park et al. \cite{park2016fingerprint} propose extending Wang's work to reduce the processing time. Instead of extracting patches from the entire image, the authors indicate that it is more efficient to extract patches from the effective fingerprint area. Thus, after extracting patches with normal probability positions of segmented fingerprint regions, the authors train those patches with CNN and make the decision using a voting strategy to make the classification.  Menotti et al. \cite{menotti2015deep}evaluate the effectiveness of the proposed newly Derived CNN architecture for spoofing detection terms using \textit{SpoofNet}. Kim et.al \cite{kim2016deep} present an FPAD method based on a Deep Belief Network (DBN) with a series of constrained Boltzmann machines connected, which learns features from training samples and determines the liveness of fingerprints. The DBN is trained in two steps. Firstly, it is trained on a set of examples without supervision from the first layer to the penultimate layer to learn the reconstruction probabilistically of its inputs. Then the model will be further trained using labeled data to perform the classification.

With an increasing number of publicly datasets set released, Chugh et.al \cite{chugh2017fingerprint} utilize Inception-v3 CNN \cite{szegedy2016rethinking} model implemented using TF-Slim library with a replacement from multiple-class softmax layer to two-unit layer for the two-class problem. Then, the model is trained with local patches extracted around the fingerprint minutiae since local patches around these minutiae are able to provide significant cues to distinguish a spoof fingerprint from live fingerprints. Along with score-level average fusion, this method evaluates several experiments such as intra-sensors and the same material, intra-sensor and cross-material, cross-sensor, and cross-dataset scenarios to consider both known and unknown attacks, which demonstrated a good average classification error (ACE). In the next work, Chugh et al. \cite{chugh2018fingerprint} further presented a Fingerprint Spoof Buster based on  the MobileNet-v1 \cite{howard2017mobilenets} model trained using  local patches that are  centered and aligned around minutia points, and defined a global \textit{Spoofness Score} to integrate the local Spoofness Score to determine the PA.

However, FPAD methods based on CNN suffer from generalization and high computational costs. The selection of PA materials used in training (known PAs) directly affects the performance against unknown PAs. Some materials (e.g. EcoFlex) has been reported to be easier to detect than others (e.g. Silgum) \cite{chugh2018fingerprint}. Hence, to further investigate the better representation of different PA materials, a new dataset namely MSU-FPAD v2.0 which combines the MSU-FPAD v1.0 and Precise Biometrics Spoof kit \cite{chugh2018fingerprint} was presented in \cite{chugh2019fingerprint}. Specifically, the database is constructed using 12 different PAIs. Then, the leave-one-out protocol is adopted; one PAI is excluded from the training set at every iteration and utilizes the 3D t-SNE technique to visualize the characteristics. Through experiments, Silicone, 2D Paper, Play Doh, Gelatin, Latex Body Paint, and Monster Liquid Latex are observed to cover the entire feature space around the Bona fide. Furthermore, it is considered a challenge to integrate existing deep learning-based algorithms with millions of parameters into an embedded or mobile device. Nguyen et.al \cite{nguyen2018fpadnet} present the FPAD technique following the architecture of the Fire module of SqueezeNet \cite{Squeezenet} and introduce the Gram Matrix \cite{Gram_matrix} to form the structure of the basis of the proposed fPADnet. This model only contains 0.3 million parameters, which is 2.4 times smaller than that of the original SqueezeNet.
Similarly, Park et al. \cite{park2019presentation} introduced a fully convolutional neural network that uses the fire module of SqueezeNet as the foundation. The model can interfere with images of any size since it has no fully connected layer. The model takes the patch extracted from the input image as input and outputs three values that show the probabilities of the classes (live, false, and background) to which the patch belongs.  Additional experiments with different input patch sizes at $32\times 32$, $48\times 48$, where $32\times 32$ and $64\times 64$ are evaluated to identify the optimum patch size to achieve the highest detection accuracy.

\begin{longtable}{c|c|p{8.355em}|p{10.355em}|p{12.355em}}
\hline
\hline
\centering

 Author & Year & Backbone & Loss function & Description \\
 \hline
 \hline
 Nogueira et.al \cite{nogueira2014evaluating} & 2014 &CNN &SVM& Extract the deep features using CNN and trained the SVM to classify the liveness\\
 \hline
Wang et.al \cite{wang2015dcnn} & 2015 &CNN&Binary CE loss & Patch based method with voting strategy\\
\hline
Menotti et.al \cite{menotti2015deep}& 2015 & CNN& Binary CE loss & Propose a Derived CNN terms as \textit{SpoofNet}\\
\hline
Kim et.al \cite{kim2016deep} & 2016 & Deep Belief Network (DBN)& Mean squared error loss&Propose a DBN based method for liveness detection\\
\hline
Park et.al \cite{park2016fingerprint} & 2016 & CNN&Binary CE loss  & Extract the patches from the effective fingerprint area to train the CNN model\\
\hline
Lazimul and Binoy \cite{lazimul2017fingerprint} & 2017 & CNN &Binary CE loss & Enhance the fingerprint image through several steps and train with CNN model\\
\hline
Jang et.al \cite{jang2017fingerprint}  & 2017 &CNN &Binary CE loss &Utilize the histogram equalization for contrast enhancement and train the CNN model with divided blocks\\
\hline
Chugh et.al \cite{chugh2017fingerprint} & 2017 &Inception-v3 CNN&Binary CE loss &Use local patches extracted around fingerprint minutiae to train the Inception-v3 model with a slightly modification of last layer\\
\hline
Chugh et.al \cite{chugh2018fingerprint}  & 2018 & MobileNet-v1&Binary CE loss& Define a global \textit{Spoofness Score} to integrate the local patch based on minutia points to determine the liveness\\
\hline
Pala \cite{pala2017deep} &2017 &CNN & Triples loss & A triplet convolutional network is proposed\\
\hline
Jung and Heo \cite{jung2018fingerprint} &2018 &CNN & Squared Regression Error(SRE) &Employ SRE layer instead of CE loss layer as loss function\\
\hline
Nguyen et.al \cite{nguyen2018fpadnet} &2018&SqueezeNet & Binary CE loss &  A lightweight model which optimize the SqueezeNet using designed Gram-Matrix \\
\hline
Chugh and Jain \cite{chugh2019fingerprint} & 2019 &CNN& Binary CE loss&Adopte leave-one-out protocol to exclude one PAI out of 12 from the training set every iteration to visualize the characteristics and find out which PAI achieves similar feature space as bona fide \\
\hline
Park et.al \cite{park2019presentation} & 2019 & SqueezeNet&Three class CE loss&A Tiny
Fully Convolutional Network without fully connected layer which can fit any size of input\\
\hline

Yuan et.al \cite{yuan2019fingerprint} & 2019 & CNN&Binary CE loss&Between the last convolutional layer and the fully connected layer, an extra Image Scale Equalization(ISE) layer is added  to preserve the texture information and maintain image resolution\\
\hline
 Jung et.al \cite{jung2019fingerprint}  & 2019&CNN& Binary CE loss&Design a Livness Map CNN (LM-CNN) to map the probe and template images to a stacked feature vector and use a Template-Probe CNN to decide the spoofness\\
 \hline
Yuan et.al \cite{yuan2019deep}& 2019 & ResNet& Binary CE loss& Introduce the ResNet with adaptive learning to tackle the gradient disappearance issue \\
\hline
 Zhang et.al \cite{zhang2019slim} & 2019&ResNet &Binary CE loss & Propose a lightweight framework that makes use of the specifically designed robust Residual block against fingerprint spoofing\\
 \hline
 Zhang et.al \cite{zhang2020fldnet} & 2020&DenseNet & Binary CE loss&Adopt the attention mechanism instead of Global
Average Pooling (GAP) \\
 \hline
 Jian et.al \cite{jian2020densely} & 2020&DenseNet& Binary CE loss& Introduce the genetic algorithm to find the
optimal DenseNet structure for fingerprint liveness detection\\
 
 \hline
 Liu et.al \cite{liu2021fingerprint}& 2021&Mobile Net V3 & Binary CE loss& Global–local model-based with rethinking strategy\\
\hline
Rai et.al \cite{rai2023mosfpad} & 2023&MoblieNet V1 & Support Vector Classifier (SVC) & Use MoblieNet V1 network for feature extraction and trained with loss from SVC\\
\hline
\caption{Existing contact based FPAD using end to end deep learning}
\label{tab:end to end DL}

\end{longtable}

Image enhancement before detection has been  well studied in deep-learning-based FPAD.  Jang et al. \cite{jang2017fingerprint} utilized histogram equalization for contrast enhancement to improve the recognition rate of fingerprint images. The fingerprint image is divided into several non-overlapped blocks and trained with a CNN model for classification. The Majority Voting System (MVS) is applied to totalize the votes of all sub-blocks and make the final decision. Similarly, Lazimul and Binoy \cite{lazimul2017fingerprint} propose to enhance a fingerprint image through six steps: Image Segmentation, Image Local Normalization, Orientation Estimation, Ridge frequency Estimation, Gabor Filtering, Image Binarization/thinning. Subsequently, a CNN model is used to train the data. Typically, cross-entropy is the most common loss function used to measure the difference between two probability distributions and is widely applied in classification tasks. Pala \cite{pala2017deep} introduces a triplet loss \cite{schroff2015facenet} that encourages dissimilar pairs to be distant from any similar pair by at least a certain margin value. A triplet network takes two patches of one class and one patch of the other to measure the inter- and intra-class distances supervised by triplet loss. Furthermore, Jung and Heo \cite{jung2018fingerprint} introduce a new CNN architecture that employs a Squared Regression Error(SRE) layer instead of a cross-entropy loss layer. This method allows setting a threshold as a liveness probability to adjust the model, which provides an accuracy trade-off option to fit different application scenarios. Furthermore, Jung et al. \cite{jung2019fingerprint} extend their previous work by introducing two CNNs terms as Livness Map CNN (LM-CNN) and Template-Probe CNN (TP-CNN). The LM-CNN performs pre-computation during fingerprint registration to map the fingerprint image to a $32 \times 32$ feature map. Then, the output map from the probe fingerprint and template fingerprint will be stacked as a $2\times 32 \times 32$ liveness map, which will be fed into the TP-CNN for the final decision. Most CNN models require a fixed length of input images because of the restriction of the fully connected layer. Thus, the fingerprint dataset requires additional preprocessing such as cropping or scaling, which leads to information loss. To address this problem, Yuan et al. \cite{yuan2019fingerprint} propose an improved DCNN model with Image Scale Equalization (ISE) to preserve texture information and maintain image resolution. Between the last convolutional layer and the fully connected layer, an extra ISE layer is added to obtain the feature map from the convolutional layer and convert the image of any scale into a fixed-length vector to fix the fully connected layer.

Furthermore, Yuan et.al \cite{yuan2019deep} first introduced the Deep Residual
Network \cite{ResNet} for FPAD. The authors designed a novel ROI extraction technique to remove the noise caused by background noise. Then, the gradient disappearance in the DCNN and learning parameters falling into local optimal value issues are tackled by applying a Deep Residual Network with adaptive learning. Owing to the concern of potential low generalization capability against unknown attack detection, a texture enhancement based on a Local Gradient Pattern (LGP) is introduced to highlight the difference between a bona fide sample and an attack sample to achieve a better generalization. 
Zhang et.al \cite{zhang2019slim} proposed a lightweight framework that makes use of the specifically designed robust Residual block \cite{ResNet} against fingerprint spoofing. Silm-ResCNN consists of nine modified residual blocks. The authors make some changes by inserting a dropout layer into each pair of convolutional layers and removing the activation function (ReLU) of the second convolutional kernel to make the model more generalized. In another specific type, the $1 \times 1$ convolutional layer is replaced with max pooling, along with a padding zero channel. Therefore, the overall structure of The Slim-ResCNN consists of Conv1, Conv2, Conv3 (Conv3$\_$1, Conv3$\_$2), and Conv4 (Conv4$\_$1, Conv4$\_$2), followed by global average pooling (Avg$\_$Pool) and a final classification layer. The model will take the extracted local patches as input and the cross-entropy
is used as the loss function. It should be noted that this method achieves the top performance in the Fingerprint Liveness Detection Competition 2017 \cite{livdet2017}, with an overall accuracy of 95.25\%. Zhang et.al \cite{zhang2020fldnet} discussed the limitation of Global Average Pooling against fingerprint spoofing and overcome it by adopting the attention mechanism. A lightweight model with only 0.48 million parameters was designed. Its blocks are designed where the residual path and densely connected path are incorporated, which benefits from DenseNet \cite{DenseNet} and ResNet \cite{ResNet}. Jian et al. \cite{jian2020densely} point out the limitations of DenseNet based architecture \cite{zhang2020fldnet} and optimize the model by adopting the genetic algorithm \cite{xie2017genetic}. Liu et.al \cite{liu2021fingerprint}
proposed a framework based on the rethinking strategy. The model consists of three modules, a global PAD module, a rethinking module, and a local PAD module. Firstly, the global PAD module receives the entire image as input and then predicts the global spoofness score. The rethinking module then takes the activation map to highlight the important regions for PAD through class activation mapping(CAM). Finally, these regions will be cropped and passed into the local PAD module to refine the prediction of the global PAD module. Recently, Rai et.al \cite{rai2023mosfpad} adopt MoblieNet V1 as a feature extractor due to the capacity of utilizing depth-wise separable
convolution operation instead of traditional convolution operation, then the network is trained by the loss obtained from SVC. Furthermore, a comprehensive comparison among many existing approaches indicates that the proposed method, namely MoSFPAD, achieved state-of-the-art results.  

\subsubsection{FPAD using transfer learning/fine tuning}

However, end-to-end deep learning-based FPAD achieved a notable improvement in classification accuracy. The size of the public fingerprint training set is insufficient to optimize a CNN model, which typically requires a large number of samples for training. On one hand, many researchers include data augmentation to apply small variations to the original data to extend the dataset. On the other hand, transfer learning and fine-tuning are normal ways to tackle the issue of small datasets. As shown in figure \ref{fig:DL contact}, transfer learning/fine-tuning is a technique that does not train a deep learning model from scratch. Instead, transfer learning uses the representations learned by a pre-trained model to extract meaningful features and outperform classification with a new classifier. Differently, fine tuning technique is to unfreeze the weights corresponding to the top few layers of a pre-trained model based on general sets of images and "fine-tune" the higher-order feature to make them more relevant for the specific task. Table \ref{tab:end to end DL performance} presents a quick overview of the existing transfer-learning-based FPAD.

Nogueira et al. \cite{nogueira2016fingerprint} extend their work \cite{nogueira2014evaluating} by utilizing transfer learning on a pre-trained VGG \cite{VGG} and AlexNet \cite{NIPS2012_4824} model and fine-tuned it using a fingerprint liveness detection dataset. By comparing four different models (two CNN models pre-trained with natural images and fine-tuned with fingerprint images, one CNN-Random model that uses only random filter weights drawn from a Gaussian distribution, and a traditional LBP-SVM model), the authors elaborate on the superiority of pre-trained CNNs on FPAD fields. Furthermore, Toosi et al. \cite{toosi2017cnn} extract a set of small-sized patches that contain foreground pixels only, and pass those patches into a pre-trained AlexNet \cite{NIPS2012_4824} with a further training step that exploits features from fingerprint datasets. Similarly, Toosi et al. \cite{toosi2017assessing} extract small-sized foreground patches of raw images and fine-tuned pre-trained AlexNet \cite{NIPS2012_4824} and VGG19 \cite{VGG} models.
Similarly, Ametefe et al. \cite{s2021fingerprint} utilize transfer learning using DenseNet(DenseNet201) \cite{DenseNet} which also achieves promising results compared with VGG and AlexNet features. 

\begin{table*}[htbp]
\centering
\begin{tabular}{c|c|p{11.355em}|p{8.355em}|p{12.355em}}
\hline
 Author& Year & Backbone & Loss function & Description \\
 \hline
 Nogueira et.al \cite{nogueira2016fingerprint} & 2016 & AlexNet, VGG & Binary CE loss   & Utilize pre-trained model and fine-tuned using fingerprint liveness detection dataset\\
\hline
Toosi et.al \cite{toosi2017cnn} & 2017 &AlexNet&Binary CE loss& Extract a set of small-sized patches containing only foreground pixels to fine-tune a pre-trained model\\
\hline
Toosi et.al \cite{toosi2017assessing} & 2017 & AlexNet, VGG19 & Binary CE loss & Utilize transfer learning on two pre-trained CNN models \\
\hline

 Ametefe et.al \cite{s2021fingerprint}& 2021&DenseNet &Binary CE loss &Utilize deep transfer learning on densenet201 network\\
 \hline
\end{tabular}
\caption{Existing contact based FPAD methods using transfer learning/fine-tuning}
\label{tab:end to end DL performance}
\end{table*}

\subsubsection{Generalized deep learning}
To improve the generalization capacity of the model, many researchers have considered applying a generalized model to transfer one domain to another using an adversarial learning-based model. The goal here is to address the generalizability of the FPAD techniques by performing domain transformation between the PAI and bona fide samples. Table \ref{tab:Generalization DL performance} presents a quick overview of existing generalizable FPAD techniques.

Pereira et.al \cite{pereira2020robust} proposed a novel model based on adversarial
training, which consists of three subnetworks: (i) an encoder network that maps the input image into a latent space,  (ii) a task-classifier network that maps the latent representation to the corresponding attack and bona fide probabilities, (iii) a species-classifier network that aims to predict the PAI species according to the attack latent representation. The species classifier is trained to minimize classification loss among the PAI species, whereas the task classifier and encoder are trained to minimize the classification loss between attacks and bona fide samples while trying to keep the PAI species classification close to random guessing. To further improve the generalization performance of the detector against spoofs made from materials that were not seen during training.  A style-transfer-based wrapper, namely, a Universal Material Generator (UMG) is proposed to reliably detect the FPAD \cite{chugh2020fingerprint}. The UMG is able to generate synthetic spoof images corresponding to unknown spoof materials by transferring the style (texture) characteristics between
fingerprint images of known spoofing materials. Then, the synthesized images provide the model with more generalization capability to detect spoofs made of unknown materials.  Sandouka et al. \cite{sandouka2021unified} propose a Unified Generative Adversarial Networks(UGAN) that can translate a single generator learning mapping across multiple domains. Subsequently, a share-weighted fusion layer acts as a classifier that fuses the output from all translated domains to determine the detection result. Similarly, Sandouka et al. \cite{sandouka2021transformers} further utilize a CycleGAN \cite{zhu2017unpaired} network for domain adoption, which transforms the source domain to the target domain. In contrast to their previous work, a transformer model is employed to take a sequence of patches of the feature map as the input. The outputs are then concatenated and projected linearly to obtain a final output that is further fed into a fully connected layer for the classification task. This work further improved the performance rather than \cite{sandouka2021unified}. Furthermore, Lee et al. \cite{lee2022towards} proposed a generalization improvement method that utilizes style transfer to transfer the material styles between fingerprints. Liu et al. \cite{liu2022fingerprint} recently proposed a channel-wise feature denoising model. They extract the 'noise' channels from the feature map by evaluating each channel of the image. Then, the interference from those channels is suppressed, and a PA-adaptation loss is designed to align the feature domain of the fingerprint. This method achieves promising results on the LivDet 2017 \cite{livdet2017} dataset.

\begin{table*}[htbp]
\centering
\begin{tabular}{c|c|p{11.355em}|p{8.355em}|p{12.355em}}
\hline
 Author& Year & Backbone & Loss function & Description \\
 \hline
 Pereira et.al \cite{pereira2020robust} & 2020&  Species-invariant neural network &Adversarial loss, species-transfer loss&An adversarial
learning approach to improve the capacity to detect the unseen attack\\
 \hline
 Chugh and Jain \cite{chugh2020fingerprint}& 2020&Universal Material Generator (UMG) & Adversarial loss, style loss, content loss& A style transfer based approach that transfer texture characteristics between fingerprint images of known materials to unknown materials \\
 \hline
 
 Sandouka et.al \cite{sandouka2021unified} & 2021 &Generative Adversarial Networks, EfficientNet V2 \cite{tan2021efficientnetv2} &Adversarial loss, reconstruction
loss, domain classification loss &A Unified Generative Adversarial Networks(UGAN) that can translate a single generator learning the mapping across multiple domains\\
 \hline
Sandouka et.al \cite{sandouka2021transformers}& 2021 &Transformers and CycleGAN&Adversarial loss, cycle consistency loss & Utilize a CycleGAN to transform from the source domain to the target domain\\
 \hline
Lee et.al \cite{lee2022towards} & 2022 &CycleGAN and CNN &Adversarial loss, binary CE loss&Transform fingerprint image from an untrained material style to a trained material
style using CycleGAN\\
\hline
Liu et.al \cite{liu2022fingerprint} & 2022&MoblieNet V2& PA-Adaptation loss, binary CE loss& Propose a channel-wise feature denoising model\\
\hline

\end{tabular}
\caption{Existing Generalized deep learning FPAD methods}
\label{tab:Generalization DL performance}
\end{table*}

\subsection{Contactless based FPAD}
\begin{figure}[htbp!]
\begin{center}
\includegraphics[width=0.9\linewidth]{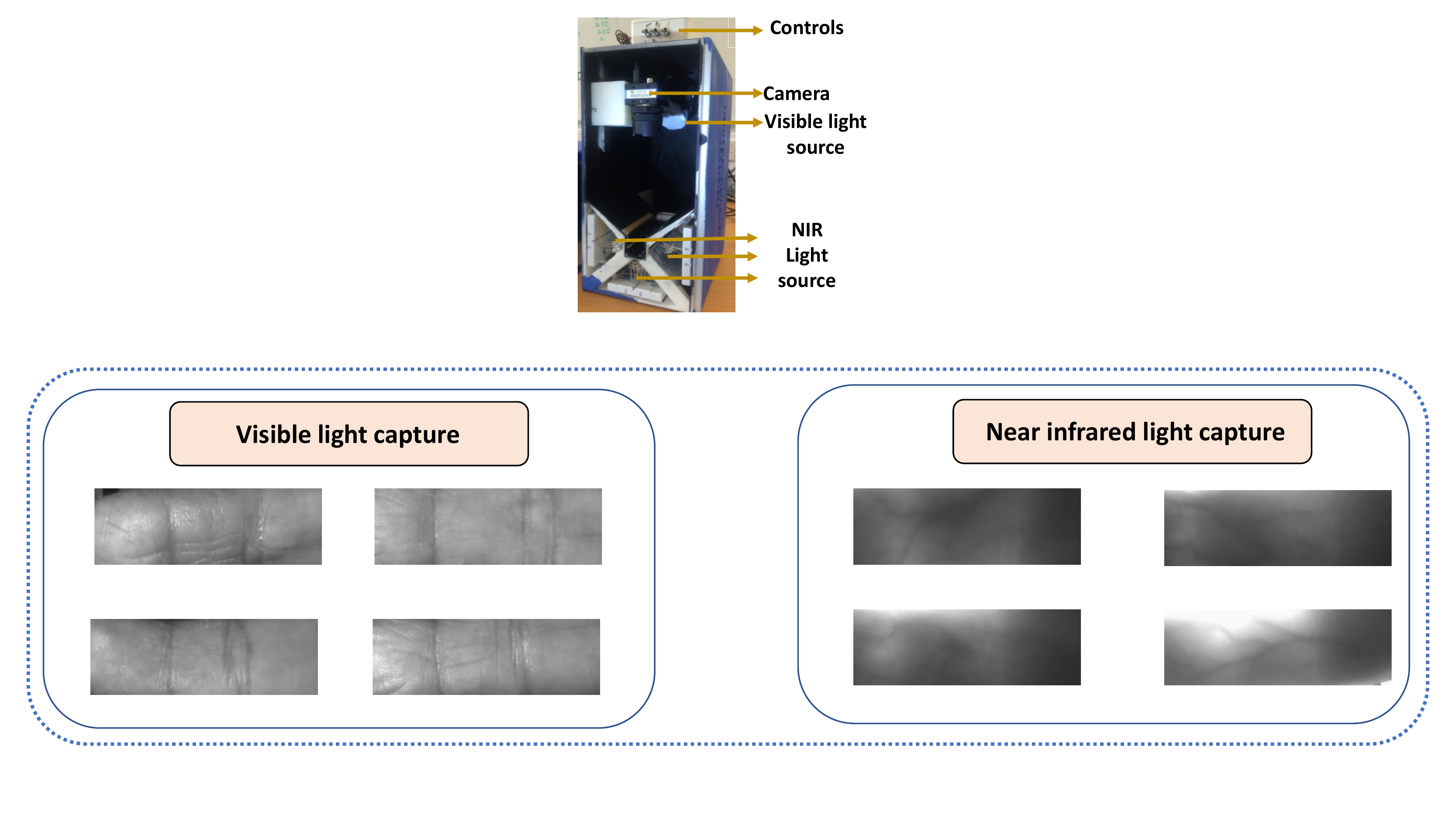}
\caption{Examples of contactless based fingerprint system using sensor-specific approaches}
\label{fig:contactless example}
\end{center}
\end{figure}

Traditional fingerprint images suffer from presentation attacks since the texture features of spoofing are typically inconspicuous. Additionally, the development of new synthetic materials brings more challenges to the generalization ability of the current models. However, the texture of the spoofing fingerprint surface can be visible through a multi-spectrum capture device or even a smartphone camera. A demonstration of multi-spectrum capture device is shown in Figure \ref{fig:contactless example}. Hussein et.al \cite{hussein2018fingerprint} propose a patch-based CNN model that takes multi-spectral short-wave infrared (SWIR) imaging and laser speckle contrast imaging (LSCI) images as input to classify the skin or no skin. Furthermore, Mirzaalian et al. \cite{mirzaalian2019effectiveness} utilize LSCI images and evaluate four different models: the model proposed by \cite{hussein2018fingerprint} terms as baseN, adding residual connections between every two 2D convolution layers of the BaseN terms as ResN, introducing inception module terms as IncpN, and a double-layer long short-term memory (LSTM)-based network. The obtained result shows that LSTM based approach achieves the best performance. 
Kolberg et al. \cite{kolberg2020analysing} analyze the long short-term memory (LSTM) network in comparison with different CNN models on a fingerprint image captured by a 1310 nm laser device. The obtained experimental results indicate the advancement of the LSTM model compared with the CNN models. Furthermore,  Spinoulas et al. \cite{spinoulas2021multi} evaluate FPAD performance under different sensing modalities using a fully convolutional neural network (FCN).  The evaluation experiments are carried under fingerprint images captured from 
Visible (VIS), near-infrared (NIR), SWIR, LSCI and Near-infrared back-illumination domains.

\begin{table*}[htbp]
\centering
\begin{tabular}{c|c|p{11.355em}|p{8.355em}|p{12.355em}}
\hline
 Author& Year & Backbone & Loss function & Description \\
\hline
Hussein et.al \cite{hussein2018fingerprint} & 2016& CNN & Binary CE loss&A patch-based CNN based on SWIR and LSCI images \\
\hline
Mirzaalian et.al \cite{mirzaalian2019effectiveness} &2019& CNN & Binary CE loss & Compare four different models \\
\hline
Kolberg et.al \cite{kolberg2020analysing} & 2020 & Long short-term memory (LSTM) network and CNN & Binary CE loss& Compare the performance among LSTM, LRCN (long-term recurrent convolutional network) and CNN models \\
\hline
Spinoulas et.al \cite{spinoulas2021multi} & 2021&fully-convolutional neural network (FCN) & Binary CE loss& Evaluate the performance with images captured from Visible (VIS), near-infrared (NIR), SWIR, LSCI and Near-infrared back-illumination \\
\hline
\end{tabular}
\caption{Existing State-Of-The-Art contactless based FPAD methods}
\label{tab:end to end DL performance}
\end{table*}

\subsubsection{Anomaly detection}
\begin{figure}[h!]
\begin{center}
\includegraphics[width=1.0\linewidth]{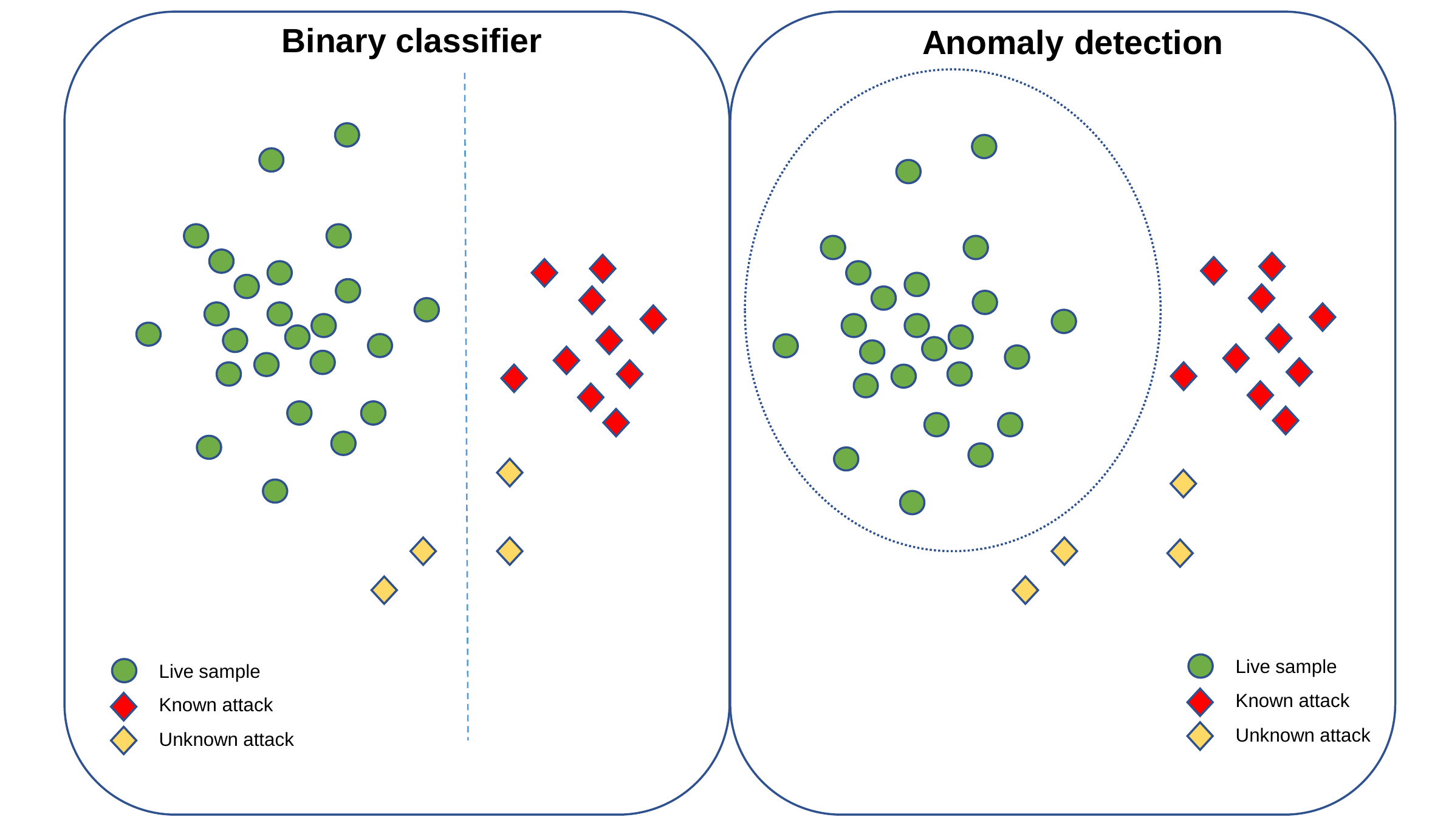}
\caption{An example of binary classifier and anomaly detection}
\label{fig:binary_vs_oneclass}
\end{center}
\end{figure}

Most previous deep learning models formulate FPAD as a close-set problem to detect various predefined PAs, which require large-scale training data to cover as many attacks as possible. In addition, the training data must be labeled prior to training. However, this leads to an overfitting issue in that the model is highly sensitive to the PAs already included in the training dataset but lacks generalization capability against unseen attacks. An increasing number of novel materials have been developed to produce gummy fingers to deceive FRS easily \cite{saguy2022proactive}. The unknown FPAD method has become an open issue and has attracted increasing attention in recent years. Compared with the most common binary classifier, the one-class classifier only learns the representation of a live fingerprint and does not use spoof impressions of any specific material. Then, the unseen attack is detected as an anomaly, which is performed as an outlier compared with the bona fide samples. Figure \ref{fig:binary_vs_oneclass} shows the difference between the binary classifier and anomaly detection-based approaches.

\begin{table*}[htbp]
\centering
\begin{tabular}{c|c|p{10.355em}|p{8.355em}|p{12.355em}}
\hline
 Author& Year & Backbone & Loss function & Description \\
 \hline
 Engelsma and Jain \cite{engelsma2019generalizing} & 2019 & Generative Adversarial Networks (GANS)& Adversarial loss& Train three GANS respectively on raw FTIR images, processed FTIR images, and direct-view images \\
 \hline
 Rohrer and Kolberg \cite{rohrer2021gan} & 2021 &Wasserstein GAN and AutoEncoder &Reconstruction loss & Utilize pre-trained WGAN weight on AutoEncoder network\\
 \hline
 kolberg et.al \cite{kolberg2021anomaly} & 2021 &Three AutoEncoders &Reconstruction loss&Three AutoEncoder are proposed \\
\hline
Liu et.al \cite{liu2021one} & 2021 & AutoEncoder &Reconstruction loss& AutoEncoder based on TOptical Coherence Technology (OCT) images\\
\hline
\end{tabular}
\caption{Existing State-Of-The-Art anomaly detection based FPAD methods}
\label{tab:end to end DL performance}
\end{table*}
Engelsma and Jain \cite{engelsma2019generalizing} propose a fingerprint spoof detector on only the live fingerprints through training multiple
generative adversarial network (GANS) \cite{GAN} for live fingerprint images. Three GAN models were trained on raw FTIR images from RaspiReader, processed FTIR images, and direct-view images. For each GAN, the generator attempts to synthesize live fingerprint images, where the discriminator uses the generated samples as well as the true samples from the dataset to distinguish them. Thus, through long iteration training, the generator is trained to generate high-quality images where the discriminator is able to separate the liveness sample from the generated sample. After the training phase, the generator is discarded, and the discriminator can be used as an FPAD module to detect attacks. Finally, fusion of the scores output by all three discriminators constitutes the final spoofness score of an input fingerprint sample. Rohrer and Kolberg \cite{rohrer2021gan} first utilize the Wasserstein GAN \cite{arjovsky2017wasserstein} as a pretrained model trained with LivDet2021 \cite{livdet2021} Dermalog Sensor dataset from scratch. The GANs discriminator weights are transferred to the AutoEncoder's(AE) encoder, whereas the generator weights are transferred to the decoder. A convolutional
layer is added between the encoder and decoder to connect them. The AE learns to minimize reconstruction loss \cite{geron2022hands} so that the model can reconstruct the input image with minimal reconstruction error. The PA will be detected with a large reconstruction error.

Kolberg et al. \cite{kolberg2021anomaly} propose a new PAD technique based on autoencoders (AE) trained only on bona fide samples captured in the short-wave infrared domain, which converts the two-class classification problem into a one-class domain. The authors introduced three AE architectures for Conv-AE, Pooling-AE, and Dense-AE, and compared the results with other state-of-the-art one-class PAD. In addition, Liu et al. \cite{liu2021one} proposes a novel One-Class PAD (OCPAD) method for Optical Coherence Technology (OCT) images that provides an internal representation of the fingertip skin rather than a simple feature. The proposed PAD framework consists of auto-encoder network-based reference bona fide modeling and spoofness score-based PA detection.

\subsection{Smartphone based FPAD}
The rapid development of smartphone-based authentication applications has achieved high verification accuracy \cite{ramachandra2023finger}, which raised concerns about the smartphone-based system being easily spoofed. Zhang et.al \cite{zhang20162d} proposes a 2D smartphone based approach that combines the features of two local descriptors (LBP and LPQ) with deep features extracted from a CNN model. The CNN model is optimized by integrating global average pooling and batch normalization. Due to the lack of publicly available datasets, a self obtained bona fide samples and 2D attack samples made of wood glue, electrosol from PCB, or printed by special conductive ink are built. By fusing the results of the two descriptors and the CNN, the final decision can be output. Fujio et.al. \cite{fujio2018face} compare the performance of using hand-crafted based method(LBP, LPQ) and deep learning based method. The obtained results indicate that the DCNN (AlexNet) can achieve a stable accuracy when the intensity of the blurring noise increases. 
Marasco and Vurity \cite{marasco2021fingerphoto} explored detection performance by training the IIITD database using various CNN architectures (AlexNet \cite{NIPS2012_4824} and ResNet18 \cite{ResNet}). The comparison results demonstrate that AlexNet achieved a robust performance against unseen attacks. The authors further propose a novel method \cite{marasco2022deep} that explores the detection effectiveness of different CNN models on different color spaces. The raw image is converted into RGB, YCBCr, HSV, LAB, and XYZ color spaces, then the five images are then further fed into five pre-trained CNN models(AlexNet \cite{NIPS2012_4824}, DenseNet201 \cite{DenseNet}, DenseNet121, ResNet18 \cite{ResNet}, ResNet34, and MobileNet-V2\cite{howard2017mobilenets}). The Best network is selected, and the score of the five color spaces is fused to obtain the final decision. Recently, to address the lack of publicly available fingerphoto presentation attack detection datasets, Purnapatra et al.. al \cite{purnapatra2023presentation} propose a new dataset comprised of six different PAIs. The FPAD methods use the state-of-the-art CNN models DenseNet 121 and NASNet \cite{zoph2018learning} which achieve promising PAD accuracy on the proposed dataset.
 

\begin{table*}[htbp]
\centering
\begin{tabular}{c|c|p{10.355em}|p{7.355em}|p{12.355em}}
\hline
 Author& Year & Backbone & Loss function & Description \\
\hline
 
 Zhang et.al \cite{zhang20162d} & 2016 & CNN&Binary CE loss& Two handcrafted-features trained by SVM integrate with CNN.\\
\hline
Fujio et.al. \cite{fujio2018face} &2018 & AlexNet &Binary CE loss & Compared with handcrafted based method \\
\hline
Marasco and Vurity \cite{marasco2021fingerphoto}  & 2021 & AlexNet, ResNet18 &Binary CE loss & Evaluate the  different CNNs on the IIITD database with spoof data including printout and various display attacks. \\
\hline
Marasco et.al \cite{marasco2022deep}& 2022 & AlexNet, DenseNet201, DenseNet121, ResNet18, ResNet34, MobileNet-V2 & Binary CE loss & Explore the effectiveness of deep representations derived from various color spaces based on the best model of six CNN models. \\
\hline
Purnapatra et. al \cite{purnapatra2023presentation}& 2023 & DenseNet 121 and NASNet \cite{zoph2018learning} &Binary CE loss &Develop a new fingerphoto PAD dataset and utilize DenseNet and NASNet for testing detection accuracy \\
\hline
\end{tabular}
\caption{Existing smartphone based FPAD methods}
\label{tab:Smartphone DL performance}
\end{table*}

\subsection{FPAD using hybrid feature extraction methods}
The hybrid method refers to combining more than one type of feature (handcrafted features, deep features, and multi-spectrum features).  etc.) to  detect the PAIs. The hybrid features can be used with all types of fingerprint capture devices and have been demonstrated to achieve higher detection accuracy at the cost of computation. Table \ref{tab:hybrid DL performance} briefly discusses the state-of-the-art hybrid FPAD methods.

Jomaa et.al  \cite{m2020end} utilize electrocardiogram (ECG) signals as well as the deep features to jointly make the decision. Furthermore, 
Tolosana et al. \cite{tolosana2018towards} propose a multi-model-based approach that utilizes four deep features respectively extracted from 5-layer residual network, pre-trained Moblie-Net-based network, pre-trained VGG 19 based network and CNN model, and combined the deep features with a spectral signal from a short-wave infrared (SWIR) spectrum capture device to jointly determine the liveness. Gomez et.al \cite{gomez2019multi}
analysis the surface of a finger within the SWIR spectrum and inside the finger using laser speckle contrast imaging (LSCI) technology. The SWIR feature is extracted by the ResNet and VGG19 network combined with the handcrafted feature extracted from the LSCI using BSIF, HOG, and LBP descriptors. Then, the fusion of the results obtained from the above techniques determines the final liveness score. Plesh et al. \cite{plesh2019fingerprint} proposes a novel approach that combines dynamic time-series features with static features extracted by the Inception-V3 \cite{szegedy2016rethinking} CNN model. Then the final result will be obtained by the fusion of two feature sets. The experiments demonstrated that the fusion of two feature sets achieves a better performance than the individual features. Kolberg et al. \cite{kolberg2021generalisation} also include SWIR and laser techniques to analyze spoofing. In contrast to \cite{gomez2019multi}, they select a long-term recurrent convolutional network (LRCN) \cite{donahue2015long}, a pre‐trained CNN model, and an autoencoder network to independently obtain a liveness score from the laser image. Meanwhile, the CNN and AutoEncoder models process the image from the SWIR. Finally, subsequent score-fusion was applied to obtain the final score for classification.

\begin{table*}[htbp]
\centering
\begin{tabular}{c|c|p{10.355em}|p{7.355em}|p{12.355em}}
\hline
 Author& Year & Backbone & Loss function & Result \\
 \hline
 
 Tolosana et.al \cite{tolosana2018towards} & 2018 &RenNet, MobileNet, VGG19, CNN& Binary CE loss& A multi-model approach that utilizes four deep features and SWIR image features\\

\hline
Gomez et.al \cite{gomez2019multi} & 2019 &ResNet and VGG & Binary CE loss&SWIR features extracted by ResNet and VGG model and LSCI features extracted by handcrafted-based methods\\
\hline
Plesh et.al \cite{plesh2019fingerprint} & 2019 &Inception-V3 & Binary CE loss & Dynamic time-series feature with static feature extracted by Inception-V3 model  \\
\hline
Jomaa et.al  \cite{m2020end} & 2020 & Mobilenet-v2 & Binary CE loss& Electrocardiogram(ECG) features and deep features\\
\hline
Kolberg et.al \cite{kolberg2021generalisation} & 2021 & Long‐term recurrent convolutional network(LRCN), CNN and  AutoEncoder& Binary CE loss, reconstruction loss&Combine the liveness score obtained from laser image and SWIR images\\
\hline

\end{tabular}
\caption{State-of-the-art hybrid FPAD methods}
\label{tab:hybrid DL performance}
\end{table*}

\section{Performance Evaluation Metrics}
\label{sec:Evaluation}
In this section, we discuss different evaluation metrics that are widely used in the fingerprint PAD literature.  First, we present the metrics used in LivDet competitions \cite{marcialis2009first} followed by ISO/IEC using ISO/IEC  30107- 3 \cite{ISO-IEC-30107-3-PAD-metrics-170227} metrics. 

\subsection{Evaluation metrics from LivDet competitions}
since the first edition of Fingerprint Liveness Detection competition (LivDet) in 2009 \cite{marcialis2009first}, the following performance evaluation metrics are used to benchmark the performance of the FPAD algorithms: 
\begin{itemize}
    \item \textit{Frej}: Rate of failure to enroll. 
    \item \textit{Fcorrlive}: Rate of the live fingerprint to be classified correctly. 
    \item \textit{Fcorrfake}: Rate of the fake fingerprint to be classified correctly. 
    \item \textit{Ferrlive}: Rate of the live fingerprint to be misclassified. 
    \item \textit{Ferrfake}: Rate of the fake fingerprint to be misclassified. 
\end{itemize}
Additional evaluation metrics such as average classification error (ACE) are defined in \cite{chugh2017fingerprint}: 
\begin{equation}
    ACE = \frac{Fcorrlive + Fcorrfake}{2}
\end{equation}
Starting from LivDet 2021 competition, evaluation metrics are defined according to the ISO/IEC 30107–1 standard presented below.
\subsection{ISO/IEC  Metrics for PAD}
International Standard Organization (ISO/IEC 30107–1:2016) \cite{ISO30107PAD} has described the general framework to present the attack detection performance results. The ISO/IEC  30107–1 framework defined following metrics: 
\begin{itemize}
    \item \textit{Liveness Accuracy}: Rate of samples correctly classified by the PAD system.
    \item \textit{APCER (Attack Presentation Classification Error Rate)}: Rate of fake fingerprints to be misclassified. The APCER for a given presentation attack instrument species(PAIS) is defined as: 
    \begin{equation}
    APCER_{PAIS} = 1 - (\frac{1}{N_{PAIS}})\sum_{i=1}^{N_{PAIS}}RES_{i}
\end{equation}
    $N_{PAIS}$ indicates the number of attack presentations for the given PAI species, $RES_{i}$ takes the value 1 if the $i^{th}$ presentation is classified as an attack
presentation, and value 0 if classified as a bona fide presentation.
    \item \textit{BPCER (Bona fide Presentation Classification Error Rate)}: Rate of the live fingerprint to be classified correctly. BPCER shall be calculated as: 
    \begin{equation}
    BPCER = \frac{\sum_{i=1}^{N_{BF}}RES_{i}}{N_{BF}}
\end{equation}
    where $N_{BF}$ indicates the number of bona fide presentations. $RES_{i}$ takes the value 1 if the $i^{th}$
presentation is classified as an attack presentation and value 0 if classified as a bona fide presentation.
\end{itemize}

D-EER (Detection Equal Error Rate) indicates that the APCER is equal to the BPCER. Normally, D-EER is valuable which stands for the performance of a biometric system. To further evaluate the performance of the integrated system, additional metrics are also defined:
\begin{itemize}
    \item \textit{FNMR (False Non-Match Rate)}: Rate of genuine fingerprints to be classified as an impostor. 
    \item \textit{FMR (False Match Rate)}: Rate of zero-effort impostors classified as genuine. 
    \item \textit{IAPMR (Impostor Attack Presentation Match Rate)}: Rate of impostor attack presentations classified as genuine.
    \item \textit{Integrated Matching Accuracy}: Rate of samples correctly classified by the integrated system. 
\end{itemize}

\section{Future Work}
\label{sec:Future}
With the revolution of deep learning, training deep neural networks (DNN) has dominated the field of image classification and object recognition. This technique was further extended to FPAD methods, and achieved notable improvements in the detection of fabricated fingerprint replicas. However, there are still some limitations that can be considered and discussed. In this section, we introduce the major challenges of current research and future perspectives. 

\subsection{Generalization to unknown attack detection}
Normally, a deep learning-based FPAD model takes both bona fide and attack samples for training so that the classifier can distinguish liveness based on the probability score calculated corresponding to the labels. However, these methods suffer from low generalization ability against PAIs not included in the training set. Another type of method is anomaly detection, which trains a one-class classifier based on only the bona fide samples to represent the real fingerprint images better to detect anomalies that achieve acceptable results against unknown attacks to some extent. In \cite{sandouka2021unified}, the authors proposed a domain-adaptation approach that can generate mappings across sensors to reduce the distribution shift between different fingerprint representations. Based on this approach as a starting point, it is worth exploring how to make the model learn the mapping from a source domain to an unseen domain to achieve a general representation of the fake fingerprint. 
\subsection{Interpretability to fingerprint presentation attack detection}
The widely deployed AI application raised the interpretability issue of how to make the deep learning model explainable. Therefore, it is important to build 'Transparent' models that have the ability to explain why they predict what they predict. In order to provide an analysis of the interpretation of FPAD methods using visualization techniques. Techniques such as \cite{zeiler2014visualizing} \cite{selvaraju2017grad} \cite{ancona2017towards} can be used to highlight the region of an image that affects the final decision. Lie et.al \cite{liu2022fingerprint} include Grad-CAM to visualize the important regions
related to the given label. It will be interesting to consider applying visualization, especially to images captured by multi-spectrum devices.

\subsection{Light-weight models for finger photo presentation attack detection}
With the rapid development of smartphone cameras, high-resolution fingerphotos can be captured efficiently and directly from a mobile device to aid reliable biometric authentication. Because mobile devices mostly do not have a high-computation environment, a lightweight deep learning model with fewer parameters and focus on only a small region of the fingerprint images would be an optimal solution. Hence, another perspective could be the presentation attack detection of smartphone-based approaches. 
\subsection{Lack of large-scale publicly available dataset}
As deep learning has shown a significant impact on the image classification domain, research on FPAD methods has concentrated on training a large-scale neural network to detect spoofing. Because of the requirement of the large scale of both bonafide and attack sample data using a deep learning model, there have been plenty of publicly available datasets published using different capture devices and spoofing materials. However, datasets comprising a large number of samples are still lacking. Especially in contactless finger photos, there is currently a lack of datasets containing bonafide and spoofing samples. A further perspective could be to produce large-scale finger photo presentation attack datasets comprised of various presentation attack instruments materials. 
\subsection{Potential adversarial presentation attack}
An adversarial attack generates adversarial samples by purpose in order to mislead the image classification result of a machine learning model. One simple way to generate adversarial examples is to add a perturbation of some pixels so that the output image looks no different from the input image, but the classification result will be changed. Another notable observation was proposed by Casula et al. \cite{casula2021spoofs}, who produce high-quality spoofs through the snapshot picture of a smartphone to obtain the fingerprint latent. Through digital processing, the spoof was fabricated using a transparent sheet. The experiment indicated that this ScreenSpoof presents a threat at the same level as a normal presentation attack. Hence, Marrone et al. \cite{marrone2021fingerprint} investigate the feasibility of adopting an adversarial attack on a physical domain by materially realizing a fake image based on an adversarial fingerprint example. The evaluation of the attack indicates that printed adversarial images exhibit a high attack rate with multiple attacks and fairly good results with a one-shot attack. According to this, it should be considered that with adversarial examples targeting the FPAD module, it is possible to combine other digital attacks(Masterprint, morphing.  etc.) as well as adversarial perturbation to fool the FPAD system to perform more dangerous attacks on the FRS system. Hence, it is interesting to exploring countermeasures against emerging adversarial attacks. 

\section{Conclusion}
\label{sec:Conclusion}

This article provides a comprehensive overview on the  common fingerprint presentation attack instruments and presentation attack detection techniques that are widely employed in both contact and contactless fingerprint biometrics. A brief introduction to publicly available databases and relevant standards regarding FPAD evaluation metrics are presented. Then a comprehensive literature review of the existing state-of-the-art deep-learning-based FPAD methods is presented, which covers contact, contactless, and smartphone-based approaches. Finally, potential future perspectives are discussed to motivate future research. Overall, this study provides an intuitive guide to researchers interested in this area.  


\begin{acks}
This work is supported by OFFPAD project funded by the Research Council of Norway.
\end{acks}

\bibliographystyle{ACM-Reference-Format}
\bibliography{reference}

\appendix

\end{document}